\documentclass[10pt,twocolumn,letterpaper]{article}
\pdfoutput=1

\usepackage{iccv}
\usepackage{times}
\usepackage{epsfig}
\usepackage{graphicx}
\usepackage{amsmath}
\usepackage{amssymb}

\usepackage[frozencache,cachedir=.]{minted}

\usepackage[square,numbers]{natbib}
\usepackage[pagebackref=true,breaklinks=true,letterpaper=true,colorlinks,bookmarks=false]{hyperref}

\newcommand{\tech}[1]{\texttt{#1}}

\iccvfinalcopy 

\ificcvfinal\fi
\begin{document}

\title{Dataset Factory: A Toolchain For Generative Computer Vision Datasets}

\author{Daniel Kharitonov\\
Stanford University\\
Stanford, CA\\
{\tt\small dkh@cs.stanford.edu}
\and
Ryan Turner\\
Iterative, Inc.\\
San Francisco, CA\\
{\tt\small ryan@dvc.org}
}

\maketitle

\begin{abstract}
Generative AI workflows heavily rely on data-centric tasks---such as filtering samples by annotation fields, vector distances, or scores produced by custom classifiers. At the same time, computer vision datasets are quickly approaching petabyte volumes, rendering data wrangling difficult. In addition, the iterative nature of data preparation necessitates robust dataset sharing and versioning mechanisms, both of which are hard to implement ad-hoc. To solve these challenges, we propose a ``dataset factory'' approach that separates the storage and processing of samples from metadata and enables data-centric operations at scale for machine learning teams and individual researchers.
\end{abstract}

\section{Introduction}
The proliferation of visual foundation models like Stable Diffusion~\citep{rombach2022high}, Florence~\citep{yuan2021florence}, and SwinTransformer~\citep{Liu_2021_ICCV} is rapidly shifting the research focus in computer vision towards data selection and data curation tasks.
This movement to Data-Centric AI (DCAI) assumes that downstream task accuracy can be improved by feeding expressive models with more data and refined metadata (e.g., as demonstrated by~\citet{Zhai_2022_CVPR})\@. While research in DCAI remains a new and exciting field, some early conclusions point towards the importance of continuous data training, actionable data monitoring, end-to-end versioning, and tighter control for metadata~\citep{DBLP:journals/corr/abs-2112-06439}.
Meanwhile, researchers aiming to work with large-scale datasets in computer vision experience significant hurdles in implementing these ``best practices'' at scale. As an illustrative example, the prestigious DataComp competition built around the Common Crawl data features four scale tracks (12.8M, 128M, 1.28B, and 12.8B samples~\citep{gadre2023datacomp}), but in the first few months, the leaderboard for the top two tracks have only attracted submissions from Meta and the DataComp team itself~\citep{board2023}.

We will briefly highlight some of the practical difficulties of doing DCAI on generative datasets below.

\subsection{Storage and access to data}
Storage is the most immediate issue when working with large-scale computer vision tasks. For example, the full LAION-5B dataset takes up to 0.5PB in volume for data plus metadata~\citep{schuhmann2022laion5b}. Therefore, working with such a dataset is feasible only using cloud storage or a networked-attached storage cluster, preferably with minimal ETL (extract, transform, and load) operations. Most inexpensive networked storage solutions, however, feature bottlenecks in both download speed and object \tech{GET} request rates. The latter limitation is especially problematic for large datasets because each sample is typically accompanied by multiple metadata files (JSON, vector embeddings, etc.)\@. As a result, even a simple task of downloading 5.8 billion image-text pairs from the cloud may take many days to complete. 

The rate-limit bottleneck in the cloud can be mitigated by grouping samples and metadata into archives (\tech{tar}, \tech{parquet}, or compressed \tech{.npz} formats) to reduce the total number of files. The trade-off in this method is losing random access to particular samples or features inside the archives. This limitation breaks most solutions that rely on constructing a URI to pinpoint the location of every sample or attribute in the storage. 

\subsection{Dataset sharing and versioning}
Curation of a generative dataset is a multi-step process by design and typically involves stages specialized for the removal of NSFW images (i.e., adult content), detection of near-duplicates, digital privacy preservation, image clustering, and more. Each stage is usually driven by an auxiliary ML model and produces a subset (or a superset in the case of data augmentation) of the original data sources. The necessity of experimenting on these curation tasks creates the dual problems of dataset versioning (keeping track of the code versions plus the experiment outcomes) and dataset sharing (passing results of each stage to a different machine or another research team)\@.

However, most computer vision datasets are not stored in a format that allows for easy versioning and sharing. For instance, the LAION-5B dataset is often shipped as a directory tree of images and their metadata in the Apache Parquet files, so sharing a subset of LAION requires repackaging and duplicating the original images and their metadata. Some older computer vision datasets (e.g., COCO~\citep{cocodataset}) appear more flexible because they offer a common JSON-based index of references that can be amended to create the derivatives; yet such formats do not support storing code alongside data, and (de-)serialization of large generative datasets into a JSON file can be a daunting task.

\subsection{Persistent auxiliary features}
Generative computer vision datasets are often distributed with some pre-computed image attributes (e.g., NSFW scores, vector embeddings, or cosine similarity between the images and their captions)\@. However, downstream filtering stages will almost certainly require more features: for example, aesthetic scores, clustering model coordinates, chromatic identifiers, etc. In many cases, it makes sense to save these (newly computed) auxiliary attributes for later reuse. Unfortunately, storage-bound dataset formats are often ill-suited for this task because the new attributes must be repackaged back into the original containers (such as JSON files or Apache Parquet tables)---which is inconvenient and slow. 


\subsection{Data provenance and incremental updates}
In its simplest form, a DCAI workflow is a linear pipeline of the processing stages that starts in storage and terminates in a data loader. A more complex DCAI workflow is best represented as a graph where multiple data sources are linked together to form a final dataset. One good example of this arrangement is the ``bring your own data'' (BYOD) track in the DataComp competition. 

Since many DCAI stages are computationally intensive, any efficient data management system must track the \emph{provenance}~\citep{Pruyne2022,simmhan2005survey} of every sample within the DCAI graph. This satisfies separate objectives: (1)~Re-run downstream processing stages only for new (previously unseen) samples, and (2)~When changing an intermediate-stage code, model, or data source, re-run only the affected downstream stages. 


\section{The ``Dataset Factory'' approach}
The key idea behind the proposed \emph{dataset factory} (DF) architecture intended to solve the aforementioned problems is realizing a fundamental difference between metadata and raw data: Metadata forms the basis for data curation and is queried frequently; meanwhile, the actual samples are only read when computing new signals or when instantiating a dataset for the ultimate model training. Moreover, while one sample may be associated with many attributes or features, the cumulative storage requirements for metadata to store them are typically an order of magnitude less than for storage of samples.\footnote{As a practical example, the DataComp dataset at the \tech{xlarge} scale stores 450TB of data in \tech{tar} files, but only 3TB of \tech{parquet} metadata.}

This asymmetry suggests data and metadata must be treated differently. We find that a good model for working with large generative datasets is to represent them as tables with rows representing samples as pointers into storage, and with columns containing the sample attributes. 

Simultaneously, the requirement to handle very large datasets dictates a switch from in-memory solutions such as \tech{pandas} (where memory limitations have been clearly documented~\citep{McKinney2023}) to the use of database technologies for the back-end dataset representation.

The initial formation of a database table representing a dataset can be done with a hybrid-ETL process that allows a sample to be sourced directly from storage (via appropriate cloud and archive format adapters), while the metadata is read, parsed, and stored in the columnar format in the same way the traditional ETL process works.

Since the bulk of data under this approach is not moved, it allows for efficient sharing and versioning where DCAI stages can pass the (relatively compact) tables to each other. Accessing a working dataset from any stage then equates to pulling a named table. Assuming that all team members have credentials to the underlying storage systems, this approach also solves the problem of efficient collaboration within the research group. This simple idea forms the basis for the bulk of the dataset factory features we highlight in the following.

\subsection{Data selection and signal processing}
As described above, the datasets in a dataset factory are internally represented as database tables. As a result, they accept the typical parallelized filtering and analytical expressions in Python or SQL syntax. Creating a new signal (for instance, running a new embedding model) in this paradigm is equivalent to writing a user-defined function (UDF) to operate on data samples and produce new features. Some UDFs (including parsing tuples of samples and their JSON metadata, or computing embedding vectors) are fairly standard and are bundled with the factory. Other UDFs are user-supplied and can be completely custom.

\subsection{Dataset production and provenance}
A dataset factory can use any suitable analytical or vector database according to performance requirements. However, unlike a typical database which treats tables as mutable entities, DF assumes datasets to be immutable. It also associates the dataset with the data sources and the code that was used to construct it. This ensures experiment reproducibility\footnote{We define reproducibility as a property where given a deterministic model and a static dataset, the system will deliver the same result.} and provenance tracking.
Changing any dataset in the DF workflow always creates a new dataset version, and may trigger the execution of the dependent downstream stages. Operation in this manner is similar to a factory pipeline (hence the name ``dataset factory''), where each DCAI stage produces a new named dataset, and any upstream changes (in source data, metadata, or the handling instructions of models) result in a new dataset version. 

%

\section{Workflow Example}
For a practical example of a dataset factory workflow, we consider the LAION-5B dataset. Let us assume the data is stored in the cloud with images in \tech{tar} files and metadata in the compressed \tech{.npz} format, and we intend to do the usual DCAI steps: filter based on some existing (pre-computed) attributes, filter based on the output of a custom ML model, and serve the final dataset for training.

\subsection{ETL stage}
\begin{figure}[t]
\begin{center}
\includegraphics[width=0.9\linewidth]{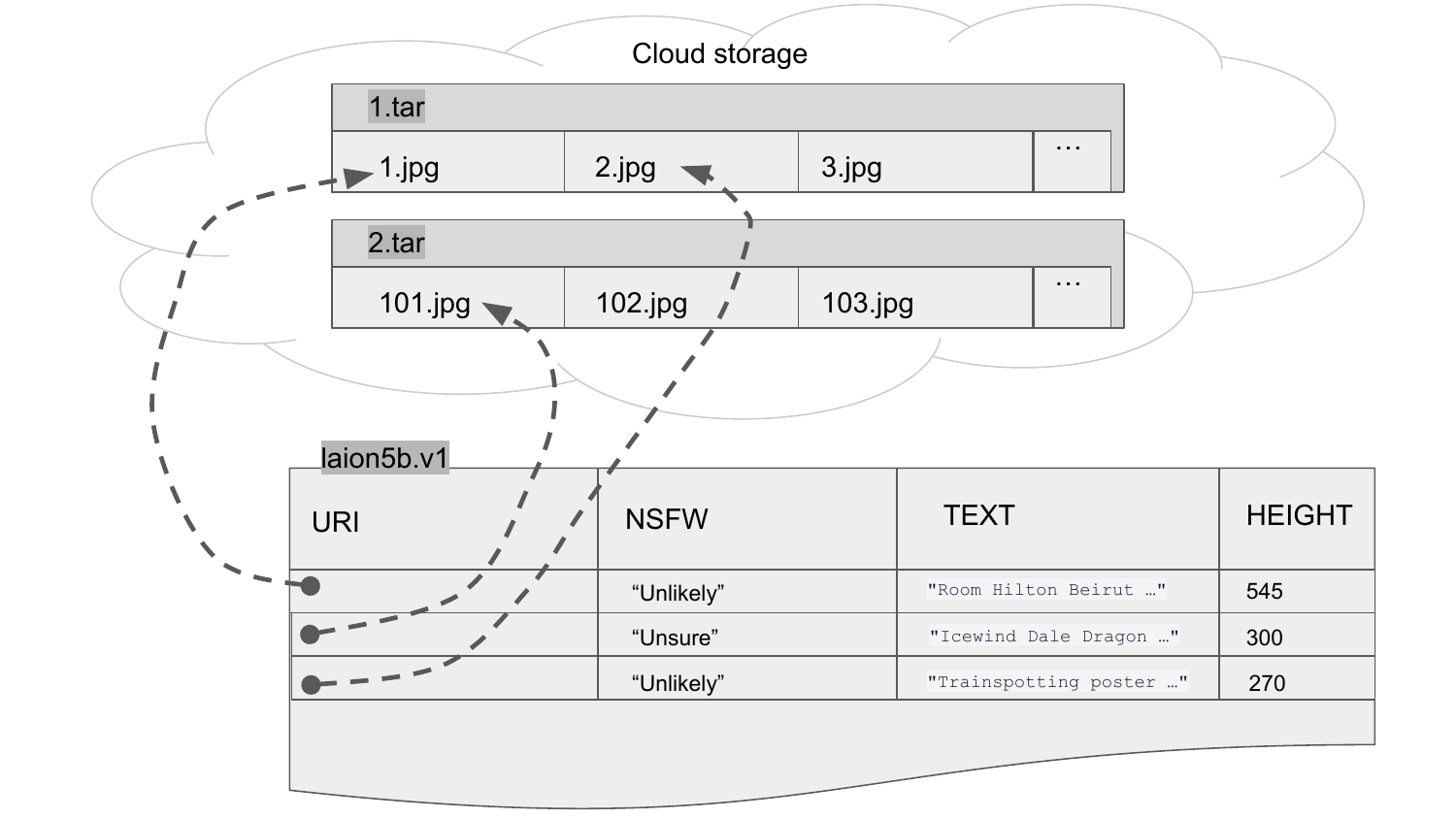}
\end{center}
   \caption{DF table representing the LAION dataset after the ETL\@.}
\label{fig:liaon5b}
\end{figure}

The first stage of the dataset factory working on LAION is to extract the attributes, match them with respective samples, and construct the schema to access the samples on-demand. Any attributes stored in separate collections are also joined in this stage. As a result, after the ETL the LAION dataset is represented as a named table \tech{laion5b.v1} seen in Figure~\ref{fig:liaon5b}. Note that references to data samples are pointing to the \tech{tar} file offsets in their respective cloud storage locations, while signals originally stored in the cloud alongside samples (not shown) were unpacked and copied into the table columns. 

\subsection{Running Queries}
Assuming we saved our table representation of LAION-5B as a named dataset, filtering by an existing attribute becomes as simple as applying a simple Boolean expression to all samples:
\begin{minted}[fontsize=\footnotesize]{python}
data_uri = "laion5b.v1"
large_images = DatasetQuery(data_uri). \
    filter(size > 1000)
\end{minted}
In the next stage, we might want to employ a custom ML model. For example, we might be interested in enriching the dataset using fashion-CLIP~\citep{Chia2022} representations.
The dataset factory API allows for a seamless addition of extra signals from a UDF purely within a Python interface that is very similar to the familiar AI coding paradigm:
\begin{minted}[fontsize=\footnotesize]{python}
fclip = FashionCLIP("fashion-clip")
large_images. \
    add_signals(fclip.encode_images)
\end{minted}
Once the embedding signal is in place, we can easily filter it down to, for example, the five hundred samples most similar to an \mintinline{python}{exemplar} image:
\begin{minted}[fontsize=\footnotesize]{python}
target, = fclip.encode_images([exemplar])

similar_images = large_images. \
    mutate(dist=cos_dist(fclip_embed, target)). \
    order_by(dist).limit(500)
\end{minted}
While the syntax for computing a new attribute remains as simple as an expression for filtering the pre-existing signal, under the hood the sequence of operations is different. An embedding ML model requires access to original samples, and the dataset factory takes care of extracting the data from storage, passing it to the embedder, and saving the results back into the dataset:
\begin{minted}[fontsize=\footnotesize]{python}
ds_uri = similar_images.save("most-similar")
\end{minted}
After this last operation, the backend database will feature a table with an additional column denoting the fashion-CLIP vector (Figure~\ref{fig:mutated})\@. Saving the dataset object as a new database table enables persistent and shared access for intermediate results of any processing stage:
\begin{minted}[fontsize=\footnotesize]{python}
very_large_images = DatasetQuery(ds_uri). \
    filter(size > 5000)
\end{minted}

\begin{figure}[t]
\begin{center}
\includegraphics[width=0.9\linewidth]{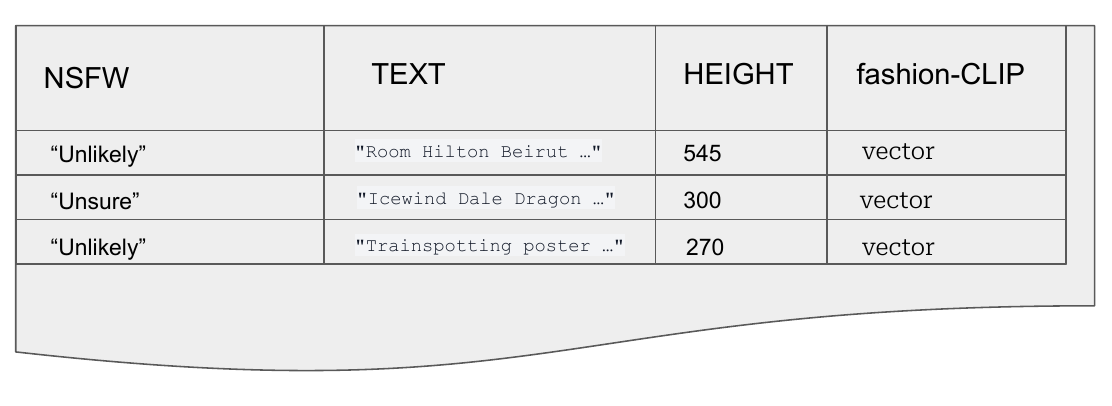}
\end{center}
   \caption{Dataset columns after adding a signal (fragment)\@.}
\label{fig:mutated}
\end{figure}

\section{Passing data to the training stage}
\begin{figure}[t]
\begin{center}
   \includegraphics[width=0.8\linewidth]{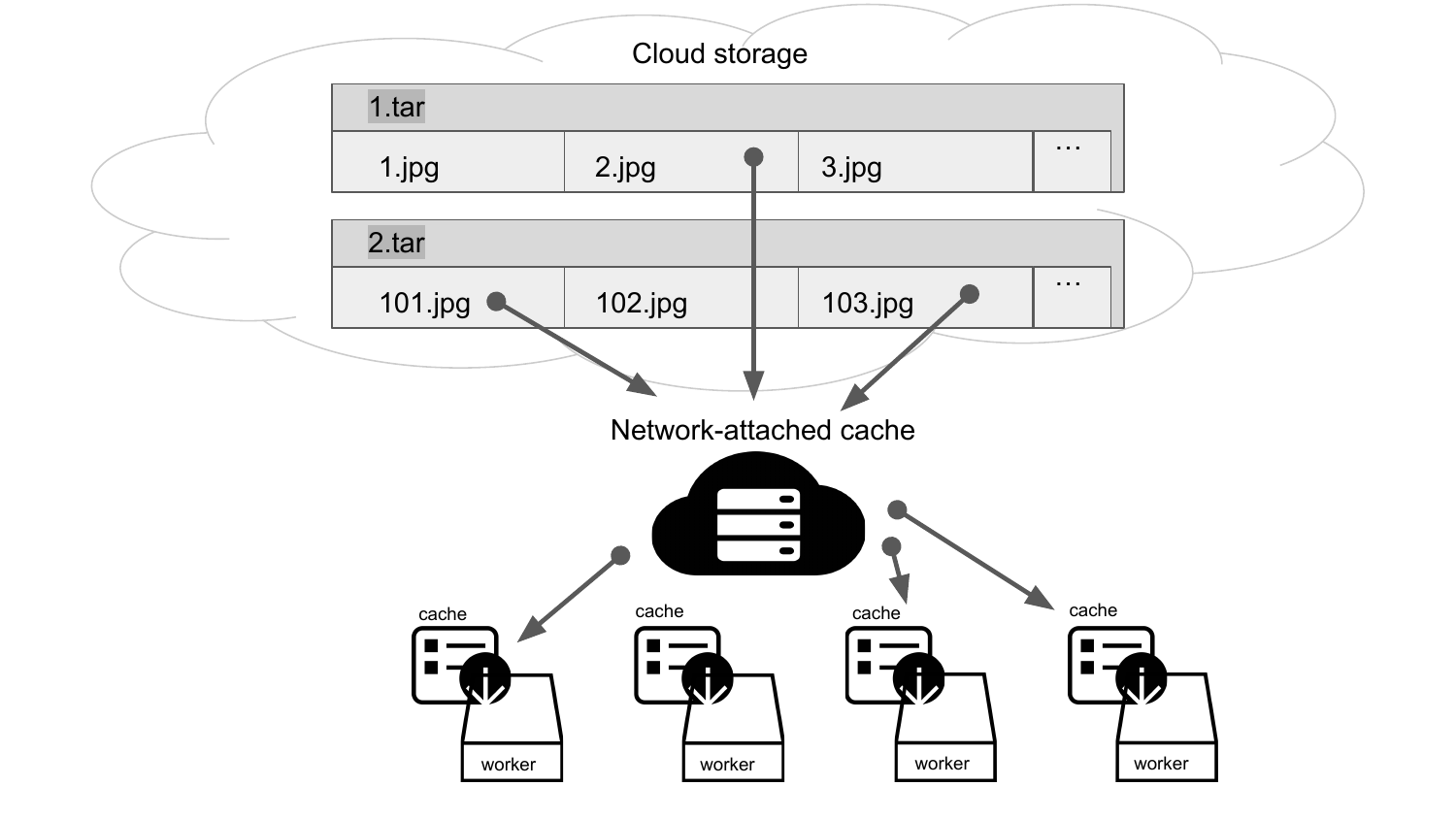}
\end{center}
   \caption{Local and intermediate caching for sample downloads.}
\label{fig:caching}
\end{figure}

So far we have described the dataset factory operation as a hierarchy of named datasets separated by the code executions: \tech{ETL $\rightarrow$ filter 1 $\rightarrow$ enrichment 2 $\rightarrow$ filter 3}, and so on. The final stage of the dataset factory terminates in a data loader object that can be passed directly to popular AI frameworks and supports distributed training. As usual, the dataset factory data loader hides the fact that the actual samples reside in the cloud, and performs the necessary data extraction steps. To speed up the repeated access to samples, the data loader also employs the local and (optionally) an intermediate cache (Figure~\ref{fig:caching})\@.

\section{Conclusions}
The dataset factory is a data representation and access approach designed for DCAI and architected to address the issues arising from curating large generative datasets. The approach is built on separating data from metadata and employs the ``dataset as a table'' abstraction to reference data samples along with their associated features. Using this abstraction, the dataset factory effectively solves the dataset versioning and sharing problem, which enables collaborative data wrangling at a very large scale.

We are providing our implementation of the dataset factory as an open-source project
and hope it will become a useful tool for AI teams in both the industry and academia.

{\small
\bibliography{egbib}
}

\end{document}